\documentclass[runningheads]{llncs}

\usepackage[T1]{fontenc}
\usepackage{inconsolata}
\usepackage{graphicx}
\usepackage{wrapfig}
\usepackage{xcolor}
\usepackage{siunitx}
\usepackage{bm}
\usepackage{booktabs}
\usepackage{multirow}
\usepackage{array}
\usepackage{hyperref}
\usepackage[nameinlink,capitalize]{cleveref}
\usepackage{enumitem}

\begin{document}

\title{Decoder-Free Supervoxel GNN for Accurate Brain-Tumor Localization in Multi-Modal MRI}

\titlerunning{Decoder-Free Supervoxel GNN}

\author{Andrea Protani\inst{1, 2}\protect\thanks{These authors contributed equally. Corresponding author \email{andrea.protani@cern.ch}.}\and
Marc Molina Van Den Bosch\inst{1,3}\protect\footnotemark[1] \and
Lorenzo Giusti\inst{1} \and \\
Heloisa Barbosa Da Silva\inst{1,4} \and
Paolo Cacace\inst{1,5} \and
Albert Sund Aillet\inst{1} \and \\
Miguel Angel Gonzalez Ballester\inst{3, 6} \and 
Friedhelm Hummel\inst{2} \and
Luigi Serio\inst{1}}

\authorrunning{Andrea Protani et al.}

\institute{European Organization for Nuclear Research, Geneva, Switzerland \and
Dept. of Neuroscience, École Polytechnique Fédérale de Lausanne, Lausanne, Switzerland \and
BCN Medtech, Dept. of Engineering, Universitat Pompeu Fabra, Barcelona, Spain \and
Universidade de Coimbra, Coimbra, Portugal \and
Sapienza Università di Roma, Rome, Italy \and
ICREA, Barcelona, Spain}

\maketitle

\begin{abstract}

Modern vision backbones for 3D medical imaging typically process dense voxel grids through parameter-heavy encoder-decoder structures, a design that allocates a significant portion of its parameters to spatial reconstruction rather than feature learning. Our approach introduces \textbf{SVGFormer}, a decoder-free pipeline built upon a content-aware grouping stage that partitions the volume into a semantic graph of supervoxels. Its hierarchical encoder learns rich node representations by combining a patch-level Transformer with a supervoxel-level Graph Attention Network, jointly modeling fine-grained intra-region features and broader inter-regional dependencies. This design concentrates all learnable capacity on feature encoding and provides inherent, dual-scale explainability from the patch to the region level. To validate the framework's flexibility, we trained two specialized models on the BraTS dataset: one for node-level classification and one for tumor proportion regression. Both models achieved strong performance, with the classification model achieving a F1-score of $0.875$ and the regression model a \text{MAE} of $0.028$, confirming the encoder’s ability to learn discriminative and localized features. Our results establish that a graph-based, encoder-only paradigm offers an accurate and inherently interpretable alternative for 3D medical image representation.

\end{abstract}

%%%%%%%%%%%%%%%%%%%%
%%% Introduction %%%
%%%%%%%%%%%%%%%%%%%%

\section{Introduction}
The automated characterization of brain tumors from multi-modal MRI has served as a key benchmark for advancing artificial intelligence in medical imaging, offering crucial support for diagnosis, surgical planning, and treatment monitoring~\cite{bakas2018identifying}. For years, convolutional neural networks (CNNs)~\cite{ronneberger2015unetconvolutionalnetworksbiomedical}, showed remarkable performance across a wide range of vision tasks \cite{Isensee2020}. The introduction of transformers motivated a shift towards attention-based architectures to overcome the limited receptive fields of CNNs and capture long-range dependencies~\cite{hatamizadeh2021unetrtransformers3dmedical,hatamizadeh2022swinunetrswintransformers,wald2025primusenforcingattentionusage}. However, both present challenges: architectural inefficiency and limited interpretability. They rely on parameter-heavy decoders, despite evidence suggesting that stronger encoders are the primary source of performance improvements ~\cite{Ibtehaz_2020,xie2021segformersimpleefficientdesign,rahman2024emcadefficientmultiscaleconvolutional}. This architectural inefficiency is compounded by a reliance on imprecise, post-hoc explainability methods like Gradient-weighted Class Activation Mapping (Grad-CAM)~\cite{DBLP:journals/corr/SelvarajuDVCPB16},  which treats interpretability as an afterthought. Together, these challenges have motivated parallel trends towards more efficient models with lightweight decoders or even decoder-free architectures~\cite{Ni2024,kerssies2025vitsecretlyimagesegmentation,brasó2025nativesegmentationvisiontransformers}, and models with built-in, rather than post-hoc, explainability~\cite{sun2024explainableartificialintelligencemedical,hou2024selfexplainableaimedicalimage}. 

To address these challenges, we introduce \textbf{S}uper\textbf{V}oxel \textbf{G}raph Trans\textbf{former} (\textbf{SVGFormer}), a new paradigm for image analysis that transforms dense volumetric data into a structured, hierarchical graph representation. Our approach begins by partitioning an image into anatomically coherent supervoxels, which serve as the fundamental nodes in our graph. Unlike traditional graph-based methods that rely on handcrafted statistical features for each node~\cite{Khatun2024,cosma2023geometric}, SVGFormer introduces a novel, end-to-end hierarchical encoding scheme. At the finest level, a patch-level feature extractor learns deep, content-aware features directly from the raw multi-modal image patches within each supervoxel. These features are then aggregated to form a rich initial representation for each supervoxel node, which is subsequently refined by a Graph Attention Network (GAT) that explicitly models the spatial and contextual relationships between neighboring regions. We posit that this flexible graph-based backbone can learn a rich, hierarchical representation, and can be adapted to other downstream tasks, from region-based property prediction, which we benchmark here with tumor segmentation on the Brain Tumor Segmentation (BraTS) dataset, to whole-graph classification. This design offers two key advantages. First, by being decoder-free, it focuses the entire computational budget on creating a high-quality encoding. Second, it allows for inherent, multi-level interpretability studies, allowing any prediction to be traced back from graph-level interactions to the specific patch features that informed it. This work lays the foundation for a new family of graph-based models, opening avenues for future research in multi-task learning, fine-grained explainability, and novel graph-centric image data augmentation.

Our primary contributions are therefore: \emph{(i)} a novel, end-to-end methodology for 3D medical image analysis that unifies voxel-level feature learning with semantic-level graph reasoning, enabling multi-level interpretability by design; \emph{(ii)} a demonstration that accurate tumor localization can be achieved without complex decoders, shifting the paradigm from high-resolution reconstruction to efficient, region-based property prediction; and \emph{(iii)} an empirical validation of a Transformer-supervoxel architecture on the BraTS dataset, establishing its competitive performance and proving its viability as a new approach for brain tumor analysis.

%%%%%%%%%%%%%%%%%%%%%%%%%%%%%%%%%%%%
%%% Background and Related Works %%%
%%%%%%%%%%%%%%%%%%%%%%%%%%%%%%%%%%%%

\begin{figure}[t]
  \centering
  \includegraphics[width=0.9\linewidth]{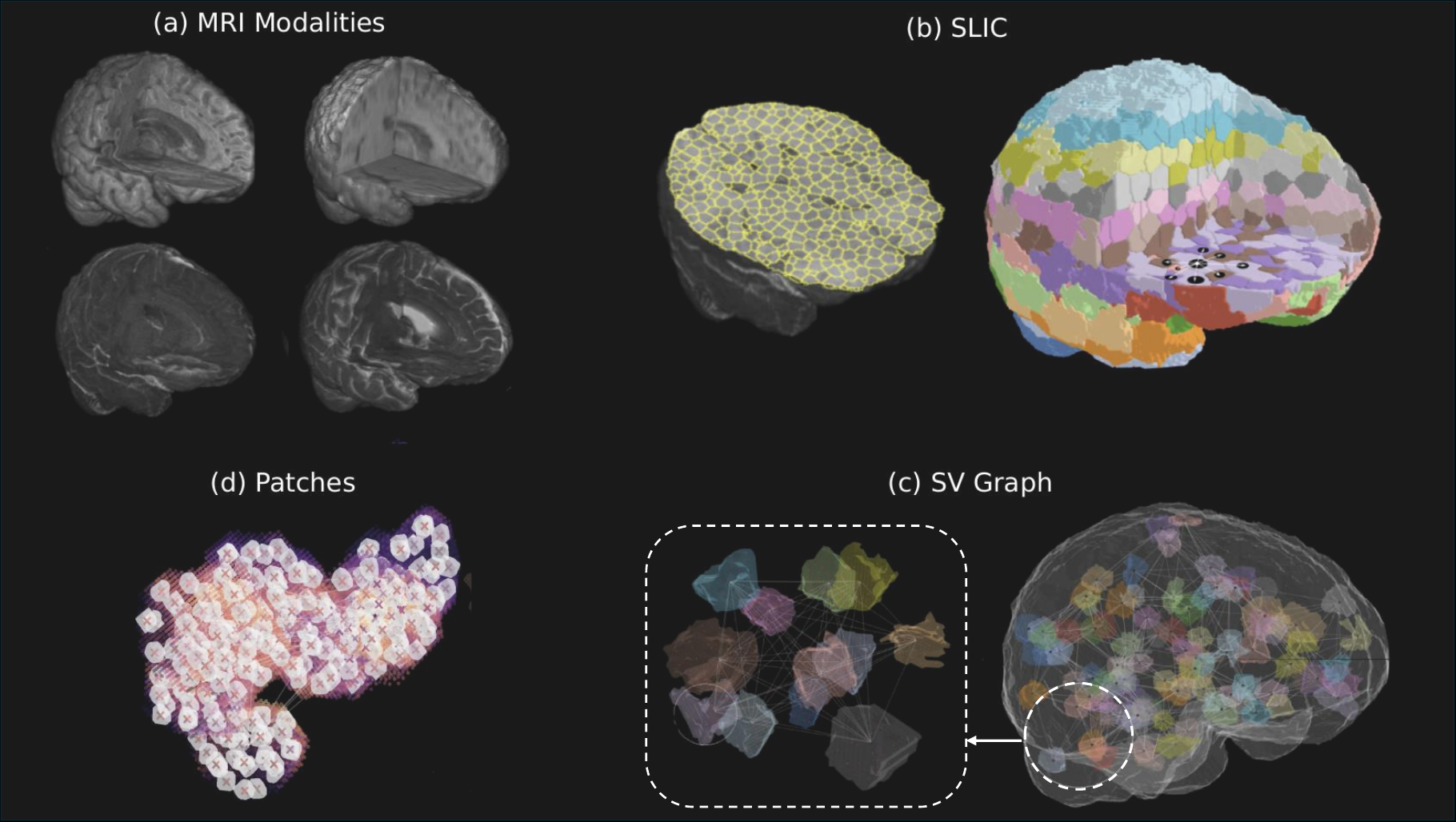}
  \caption{Pipeline for supervoxel (SV) graph 3D-image encoding: (a) Initial MRI modalities;  (b) Apply SLIC to T1-WI (SV grid shared across co-registered modalities); (c) Reconstructed graph of SVs as nodes connected to their 8 nearest neighbors; (d) Patchify method, relying on SV sampling of flattened voxel patches. While the graph is dense and packed, for visualization purposes a sparse graph is represented.}
  \vspace{-0.5cm}
  \label{fig:pipeline}
\end{figure}

\section{Background and Related Works}
\label{sec:background}

\paragraph{Deep Learning on Volumetric Brain MRI.}
Encoder–decoder CNNs such as U-Net~\cite{ronneberger2015unetconvolutionalnetworksbiomedical} and 3D U-Net~\cite{çiçek20163dunetlearningdense} remain predominant for volumetric brain MRI because their contracting path aggregates multi-scale context. Yet, the dense decoder inflates memory cost and distributes parameters across millions of up-sampling weights that do not directly improve feature extraction.

\paragraph{Transformer-based Encoders for Medical Images.}
Vision Transformers (ViT) showed that global self-attention can match or exceed convolutions on natural images~\cite{dosovitskiy2021imageworth16x16words}.  Medical adaptations (TransUNet, Swin-UNet, etc.) combine ViT encoders with lightweight decoders, reducing the parameter count while capturing long-range context~\cite{chen2021transunettransformersmakestrong,cao2021swinunetunetlikepuretransformer}.  More recent hybrids move even further toward \emph{encoder-only} designs, indicating a trend towards the investment of computational resources in feature learning rather than pixel-wise reconstruction.

\paragraph{Graph Neural Networks for Irregular Domains.}
Graph Neural Networks (GNNs) propagate information along the edges of a graph and are naturally suitable for anatomical structures~\cite{kipf2017semisupervisedclassificationgraphconvolutional}.  GAT introduces masked self-attention for learnable neighbor weighting~\cite{velivckovic2017graph}, and GATv2 refines this with dynamic attention coefficients that improve expressiveness without extra cost~\cite{brody2021attentive}. Recent work highlight their growing role in neuro-imaging for modelling long-range spatial relations and providing interpretable decision paths~\cite{luo2024graphneuralnetworksbrain}.

\paragraph{Supervoxel-Based Representations.}
Supervoxels group locally uniform voxels into anatomically meaningful regions. SLIC is the standard baseline for 2D and 3D over-segmentation~\cite{slic}, while energy-optimized variants such as 3D SEEDS \cite{bergh2013seedssuperpixelsextractedenergydriven} deliver order-of-magnitude speed-ups on large clinical datasets~\cite{zhao2025extendingseedssupervoxelalgorithm}. By transforming dense volumes into sparse region graphs, supervoxels make the application of GNNs and Attention mechanisms computationally feasible, especially in early encoding stages where the high dimensionality of the data would otherwise be prohibitive.

%%%%%%%%%%%%%%%%%%%%%%%%%%%%%%%%%%%%
%%% Supervoxel Graph Transformer %%%
%%%%%%%%%%%%%%%%%%%%%%%%%%%%%%%%%%%%

\section{Supervoxel Graph Transformer}\label{sec:method}

\paragraph{Pre-processing Pipeline.} We adopt a graph–based representation of the BraTS 2025 multi-modal 3D MRI scans.  
Each patient volume comprises four modalities: T1‐weighted (T1-WI), contrast-enhanced T1 (T1ce), T2‐weighted (T2-WI), and T2-FLAIR (FLAIR) stored at native resolution, along with a voxel-level tumor segmentation mask.  
As illustrated in~\Cref{fig:pipeline}, our pre-processing pipeline converts the raw multi-modal MRI data into a compact supervoxel-based graph structure, encoding both spatial and feature-level information. The detailed processing steps are as follows:

\begin{enumerate}[label=\roman*.]

    \item \emph{Supervoxel Generation:} To convert the continuous image domain into a graph structure, each normalized T1-WI volume is partitioned into a large number of small, locally uniform regions called supervoxels using 3D SLIC clustering. The number of supervoxels, \(n_{\mathrm{SV}}\), is a tunable hyper-parameter that controls the granularity of the graph (see~\Cref{sec:exp_setup}). The T1-WI modality is chosen for its stable anatomical contrast, and the resulting supervoxel map is applied uniformly across all modalities to ensure anatomical consistency.

    \item \emph{Dynamic Background Pruning:} To remove the background, supervoxels whose mean T1-WI intensity is below a data-driven threshold are discarded. We compute the largest gap in the sorted negative-mean distribution and set the cut-off $\theta=\frac{1}{2}\bigl(m_{(g)}+m_{(g+1)}\bigr)$ halfway between the two means on either side of that gap. All subsequent processing is restricted to the retained index set $\texttt{L}=\{\,\ell\mid\bar{I}_\ell>\theta\,\}$.

    \item \emph{Supervoxel–level tumor masks:} given the 4-class BraTS segmentation, we compute the fraction of voxels within each supervoxel belonging to any tumor class, obtaining a continuous regression target $y_{reg} \in [0, 1]$. In parallel, we define a classification ground truth by binarizing these proportions using a fixed threshold $\tau$, thus assigning each supervoxel to a binary class (tumor vs. non-tumor).

    \item \emph{Patch Extraction:} Within each retained supervoxel $\ell$, we use k-means++ to select $n_{\mathrm{patch}}$ centroids~\cite{choo2020kmeansstepsyieldconstant}. Patches are formed from the flattened $s$ nearest-neighbor voxels to each centroid, based on Euclidean distance in world coordinates. Each patch is augmented with its centroid coordinates, and features from all modalities are concatenated, resulting in a supervoxel tensor $X_\ell$ of shape ($n_{\text{patch}} \times n_{\text{modalities}}$, $s+3$).

    \item \emph{Graph Construction:} for each supervoxel node, we compute its centroid as the mean XYZ coordinate of its patches. We then link every node to its $k_{\mathrm{NN}}{=}8$ nearest neighbors and record the connections in a symmetric adjacency matrix $A\!\in\!\{0,1\}^{|\texttt{L}|\times|\texttt{L}|}$, where $A_{ij}=1$ iff nodes $i$ and $j$ are mutual $k_{\mathrm{NN}}$ neighbors.
\end{enumerate}

\begin{figure}[t]
 \centering
  \rotatebox{270}{\includegraphics[width=0.50\linewidth]{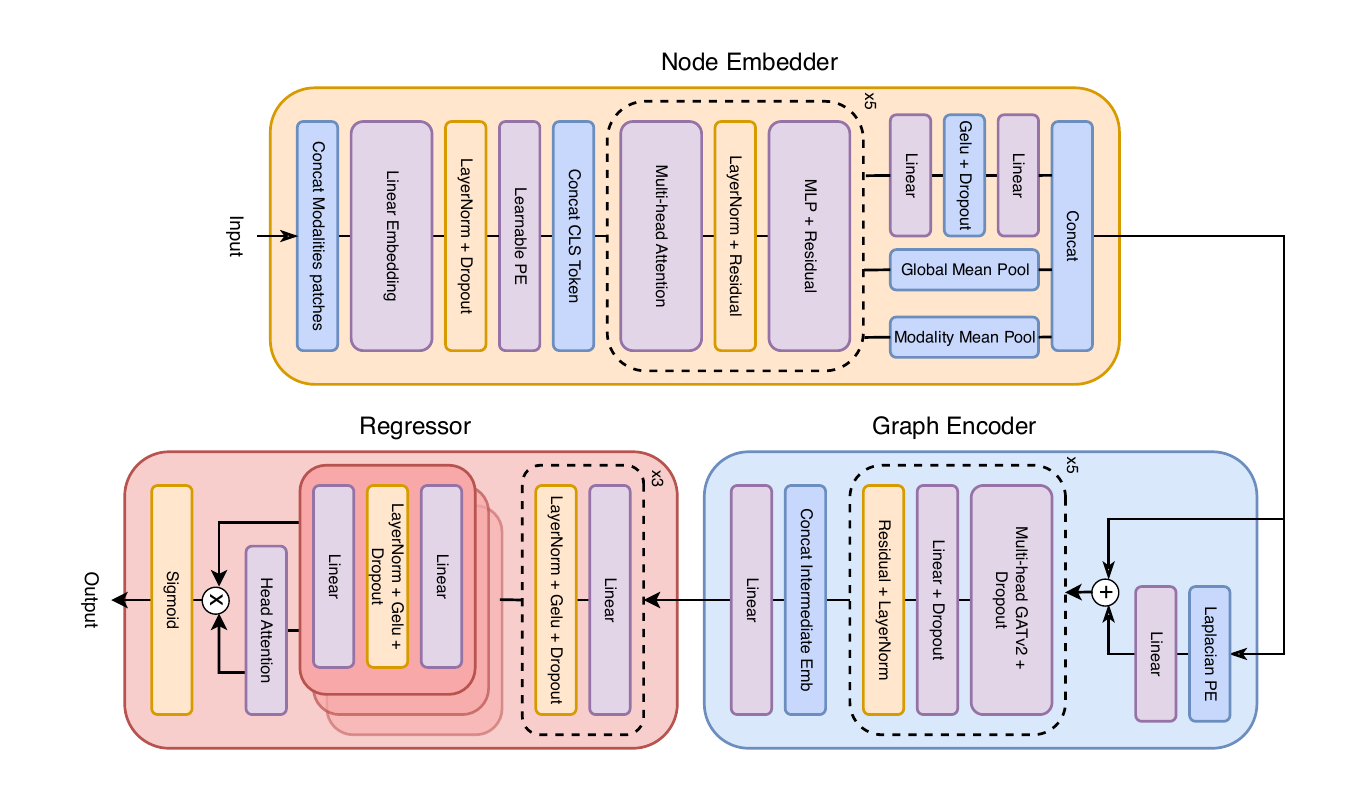}}
 \caption{Architecture of the SVGFormer; Hierarchical Encoder and Regressor.  
\emph{(i) Node Embedder}: modality-aware patch tokens pass through a 5-layer Transformer to form node features.  
\emph{(ii) Graph Encoder}: a 5-layer GATv2, augmented with Laplacian positional encodings, propagates context across supervoxels.  
\emph{(iii) Regressor}: an attention-weighted ensemble of Multi-Layer Perceptron (MLP) heads predicts the tumor fraction for each node.}
 \label{fig:architecture}
 \vspace{-0.5cm}
\end{figure}

\paragraph{Patch-Graph Neural Network Architecture.}
The SVGFormer deep-learning module is the learnable core of the proposed end-to-end pipeline and is organized into three stages. As illustrated in~\Cref{fig:architecture}, it begins with a \textbf{Node Embedder} that processes the patch tensor for each supervoxel ($X_\ell$).
Patches are linearly projected into a 256-dimensional space, summed with learnable modality embeddings, and prefixed with a \textsc{[CLS]} token. This sequence is then processed by a 5-layer, 8-head Transformer encoder. The final node representation is a rich descriptor formed by concatenating the output of the \textsc{[CLS]} token, the mean of all patch embeddings, and the per-modality means. Second, a \textbf{Graph Encoder} refines these features by modeling inter-supervoxel relationships. We employ a 5-layer GATv2, augmented with Laplacian positional encodings, to propagate context. A multiscale fusion strategy combines outputs from all GNN layers to yield a final, context-rich embedding for each supervoxel. Finally, a \textbf{Regressor} head predicts the tumor proportion. These embeddings pass through a shared MLP into an ensemble of 8 parallel prediction heads. A dedicated attention module then computes a weighted average of these outputs to produce the final sigmoid-scaled prediction in $[0, 1]$, guided by an auxiliary loss that encourages diverse head specialization.

An advantage of the SVGFormer architecture is its built-in, dual-scale explainability. High-level predictions made on a supervoxel node by the Regressor can be traced back through the Graph Encoder to identify which neighboring supervoxels were most influential. Furthermore, the feature representation of any individual supervoxel can be analyzed by examining the attention weights within its patch-level Transformer encoder. This allows a direct link from a final prediction to the specific intra-supervoxel patch features that contributed most.

\section{Experiments}\label{sec:exp_setup}

\paragraph{Dataset.}
All experiments are conducted on the Pre-Treatment Glioma BraTS 2025 training release, which contains 1251 samples. Each sample includes four co-registered MRI modalities (T1-WI, T1ce, T2-WI, and T2-FLAIR) and a voxel-level ground-truth mask delineating enhancing core, non-enhancing core, and peritumoral edema. As detailed in~\Cref{sec:method}, we normalize each volume and over-segment it into supervoxels at four distinct granularities, $n_{\text{SV}}\in$\{$1000$, $2000$, 3000, 4000\}, to analyze the effect of graph resolution. For every supervoxel node, we then construct two distinct targets to facilitate the training of separate models: (i) a continuous \emph{regression label}, $y_{\text{reg}}\in[0,1]$, corresponding to the fraction of constituent voxels belonging to any tumor class, and (ii) a binary \emph{classification label}, derived by thresholding $y_{\text{reg}}$ at $\tau=0.15$. Our evaluation protocol is based on a rigorous 5-fold cross-validation with patient-level splits to ensure that test sets are mutually exclusive and all scans from a single patient reside in the same fold to prevent data leakage. One split was reserved for an extensive hyper-parameter search.

\paragraph{Implementation Details.}
Models are implemented in PyTorch $2.2$ and PyTorch Geometric $2.5$, trained on a single NVIDIA A100 ($40$ GB) GPU.
The architecture has 39M parameters (7M embedder, 28M graph encoder, 4M predictor).
We use the AdamW optimizer (\(\text{lr}=3\!\times\!10^{-5}\), weight-decay \(=0.01\)) with cosine-annealing restarts (\(T_{0}=100\) epochs, \(\gamma=0.5\)).  
Each fold is trained for $72$ hours with a batch size of $2$ and accumulation of gradients in $8$ steps. All random seeds are fixed per fold for reproducibility.

\paragraph{Evaluation Metrics.}
For \textbf{classification} we report F1-score and ROC-AUC. For \textbf{regression} we report MAE, \(R^{2}\).  Metrics are averaged over the four evaluation folds and expressed as mean $\pm$ standard deviation.

\paragraph{Experimental Results.} To validate the flexibility of the SVGFormer encoding framework, we instantiated and trained two separate, specialized models: a \textbf{classification model} to assign a binary tumor label to each supervoxel, and a \textbf{regression model} to predict the precise tumor fraction within each supervoxel. Both models leverage the same core encoder but are optimized for their distinct tasks. We evaluated both models across four levels of graph granularity ($n_{\text{SV}}$) to assess their performance and robustness, with the results summarized in \Cref{tab:granularity}.

\begin{figure*}[t]
    \centering
    \includegraphics[width=0.9\textwidth]{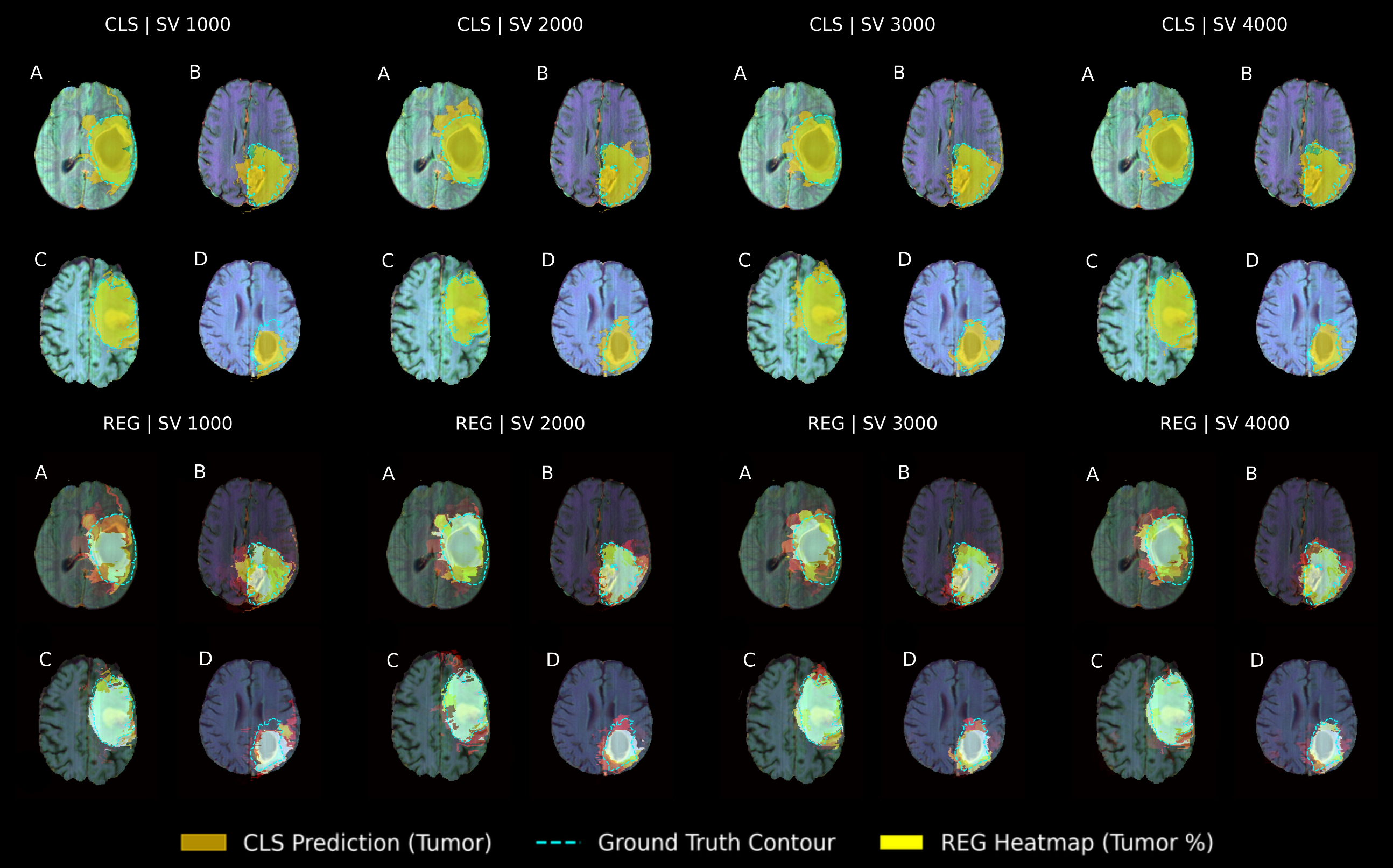}
    \caption{Qualitative analysis of model predictions on four different patients. The top row shows results from binarized \emph{Classification} models, while the bottom row shows predictions from the \emph{Regression} models. From left to right, the columns correspond to models trained with $\mathit{1000}$, $\mathit{2000}$, $\mathit{3000}$, and $\mathit{4000}$ \emph{supervoxels}. Each of these eight cells is itself a $2\times 2$ quadrant, displaying a sagittal view of the model's prediction for each of the four patients (identified as A-D at the corner).}
    \label{fig:qualitative_comparison}
    \vspace{-0.5cm}
\end{figure*}

\begin{table}[t]
\centering
\caption{Performance of two specialized models built on the SVGFormer encoder: a classification model and a regression model. Results show mean $\pm$ SD across four folds under varying graph granularities. These metrics evaluate the quality of node-level predictions for each task.}
\label{tab:granularity}

\newcolumntype{C}{>{\centering\arraybackslash}p{2.5cm}}
\newcolumntype{R}{>{\raggedleft\arraybackslash}p{2.5cm}}
\begin{tabular}{@{}l R R R R@{}}
\toprule
\multirow{2}{*}{\bfseries $n_{\text{SV}}$} &
\multicolumn{2}{c}{\bfseries Classification $\uparrow$} &
\multicolumn{2}{c}{\bfseries Regression} \\
\cmidrule(lr){2-3}\cmidrule(lr){4-5}
 & \multicolumn{1}{C}{F1-score} & \multicolumn{1}{C}{ROC-AUC} & \multicolumn{1}{C}{MAE $\downarrow$} & \multicolumn{1}{C}{$R^{2} \uparrow$} \\
\midrule
1000 & \bfseries 0.875 $\pm$ 0.006 & \bfseries 0.976 $\pm$ 0.003 & \bfseries 0.028 $\pm$ 0.001 & \bfseries 0.793 $\pm$ 0.009 \\
2000 & 0.873 $\pm$ 0.006 & \bfseries 0.976 $\pm$ 0.003 & 0.030 $\pm$ 0.001 & 0.749 $\pm$ 0.022 \\
3000 & 0.863 $\pm$ 0.013 & 0.973 $\pm$ 0.006 & 0.032 $\pm$ 0.001 & 0.718 $\pm$ 0.014 \\
4000 & 0.863 $\pm$ 0.003 & 0.973 $\pm$ 0.003 & 0.034 $\pm$ 0.001 & 0.697 $\pm$ 0.020 \\
\bottomrule
\end{tabular}
\end{table}

\paragraph{Model Performance.}
The classification model demonstrates consistently high performance across all granularities. With F1-scores remaining high (0.863--0.875) and ROC-AUC scores exceeding 0.97, the model can predict the presence or absence of the tumor in each supervoxel. This confirms that SVGFormer encoder learns discriminative features for localized binary classification objectives. Similarly, the regression model shows strong predictive capability. The low MAE, particularly at the 1000-SV level (\textbf{0.028}), and the high coefficient of determination ($R^2=0.793$) indicate that the encoder is capable of learning from a fine-grained target such as the localized tumor proportion. As granularity increases, node-level error slightly increases; this does not correspond to a clear degradation in the qualitative visualizations~\Cref{fig:qualitative_comparison}.

\paragraph{Framework Validation.}
Overall, these experiments validate our central hypothesis that a multi-stage graph-based supervoxel encoding pipeline serves as a powerful and versatile backbone for learning rich representations. It can be effectively adapted to distinct, node-level prediction tasks, both classification and regression, achieving high performance in both domains. Moreover, its applicability could be broadened by adding specific layers on top for  tasks such as dense-segmentation or image reconstruction, all relying on a shared encoder backbone.  To test the generalization power of our encoder, we evaluated its features on a downstream segmentation task without requiring any finetuning with a segmentation loss or adding any layer. Fixing a threshold over the output of the regression model slightly higher than the associated absolute error ($\tau=0.04$),  we achieve mean Dice scores ranging from $0.62 - 0.75$. Hence, this framework produces spatially coherent results for a task it was not trained on, proving robust region-specific representations.

\section{Conclusion and Future Work}

In this work, we introduced \textbf{SVGFormer}, a flexible, decoder-free framework that effectively bridges the gap between dense volumetric data and semantic graph processing for 3D medical image analysis. By converting multi-modal MRI volumes into supervoxel graphs, our end-to-end pipeline allocates its entire computational budget to a powerful Transformer-GAT encoding stack. Our experiments demonstrate that this approach produces high-quality, node-level representations for both classification and regression tasks.

The learnable graph-based image encoding pipeline serves as a backbone for general, multi-task models that could be used for finetuning specific model variations for predicting tumor severity, patient survival, and other clinical outcomes. A key advantage of this approach is its inherent multi-level explainability; the model's dual-scale attention mechanisms provide a clear path to trace high-level predictions back to the specific patch-level features that informed them, a crucial step for clinical translation. Furthermore, we plan to extend this architecture to perform fine-grained, decoder-free multi-class segmentation by regressing a vector of different tissue proportions for each supervoxel.

In conclusion, our work establishes that graph-based, encoder-only pipelines can be both performant and interpretable. By shifting the focus from voxel-level reconstruction to region-based graph encoding, SVGFormer provides a robust foundation for a new generation of multi-task, explainable models in medical imaging.

\clearpage

\bibliographystyle{abbrv}
\bibliography{bibliography}

\end{document}